\documentclass{article}

\newcommand{\onlyARXIV}[1]{#1}
\newcommand{\onlyIJCAI}[1]{}

\onlyIJCAI{%
  \usepackage{ijcai15}
  \usepackage{times}
}

\onlyARXIV{%
  \usepackage[affil-it]{authblk}
  \usepackage{fullpage}
  \newcommand{\shortcite}[1]{\cite{#1}}
}

\usepackage{xcolor,xspace}
\usepackage[normalem]{ulem} 
\usepackage{upgreek} 

\usepackage{amssymb, amsmath, amsopn, amsthm}
\usepackage{mathrsfs}
\newtheorem{theorem}{Theorem}
\newtheorem{lemma}[theorem]{Lemma}
\newtheorem{proposition}[theorem]{Proposition}
\newtheorem{definition}[theorem]{Definition}
\newtheorem{corollary}[theorem]{Corollary}
\newtheorem*{claim}{Claim}

\usepackage{dsfont}
\def\Re{\mathds{R}}
\def\Na{\mathds{N}}
\def\Z{\mathds{Z}}
\def\Co{\mathds{C}}

\newcommand{\Oh}{\mathcal{O}}

\def\E{\ensuremath{\mathrm{E}}}

\DeclareMathOperator\erfc{erfc}
\DeclareMathOperator{\poly}{poly}
\newcommand{\fnoise}[1][{\sigma^2}]{\ensuremath{\textsc{om}_{[{#1}]}}}
\def\misclass{\ensuremath{\Phi}}

\usepackage[algo2e,ruled,vlined,linesnumbered]{algorithm2e}
\SetAlgoSkip{}
\usepackage{balance}
\usepackage{url}
\usepackage{enumitem}

\usepackage{graphicx}
\usepackage{tikz}
\usetikzlibrary{positioning}
\usepackage{pgfplots}
\usepackage{pgfplotstable}
\usepgfplotslibrary{fillbetween} 

\onlyARXIV{%
  \author{Tobias Friedrich}
  \author{Timo K{\"o}tzing}
  \author{Martin Krejca}
  \author{Andrew M. Sutton}
  \affil{Friedrich-Schiller-Universit{\"a}t Jena, 07743 Jena, Germany}
}
\onlyIJCAI{%
  \author{}
}

\title{The Benefit of Sex in Noisy Evolutionary Search}

\newcommand{\ignore}[1]{}

\newcommand{\EA}[1]{($#1$)-EA}

\newcommand{\whp}{w.h.p.\xspace}
\newcommand{\om}{OneMax\xspace}

\begin{document}

\maketitle

\begin{abstract}
The benefit of sexual recombination is one of the most fundamental questions both in population genetics
and evolutionary computation. It is widely believed that recombination helps solving difficult optimization problems.
We present the first result, which rigorously proves that it is beneficial to use sexual recombination
in an uncertain environment with a noisy fitness function. 
For this, we model sexual recombination with a simple estimation of distribution algorithm called the Compact Genetic Algorithm (cGA), which we compare with the classical $\mu+1$~EA. 
For a simple noisy fitness function with additive Gaussian posterior noise $\mathcal{N}(0,\sigma^2)$,
we prove that the mutation-only $\mu+1$~EA typically cannot handle noise in polynomial time for $\sigma^2$ large enough
while the cGA runs in polynomial time as long as the population size is not too small. 
This shows that in this uncertain environment sexual recombination is provably beneficial.
We observe the same behavior in a small empirical study.
\end{abstract}


\section{Introduction}
Heuristic optimization is widely used in artificial intelligence for solving hard optimization problems,
for which no efficient problem-specific algorithm is known. Such problems are
typically very large, noisy and heavily constrained and cannot
be solved by simple textbook algorithms.
The inspiration for heuristic general-purpose  problem solvers often comes from nature. 
A well-known example is simulated annealing, which is inspired from physical annealing in metallurgy.
The largest and probably most successful class, however, are
biologically-inspired algorithms, especially \emph{evolutionary algorithms}.

\medskip
\noindent\textbf{Evolutionary and genetic algorithms.}
Evolutionary Algorithms (EAs) were introduced in the 1960s and have been
successfully applied to a wide range of complex engineering
and combinatorial problems~\cite{EibenSmith03,NeumannWitt:b:10,BackFZ97}.
Like Darwinian evolution in nature, evolutionary algorithms construct new solutions from old ones
and select the fitter ones to continue to the next iteration.
The construction of new solutions from old ones, so-called reproduction, can be \emph{asexual} (mutation of a single individual) or \emph{sexual} (crossover of two individuals).
An EA which uses sexual reproduction is typically called \emph{Genetic Algorithm} (GA).
Since the beginning of EAs, it has been argued that GAs should be more
powerful than pure EAs which use only asexual reproduction~\cite{Goldberg1989}.
This was debated for decades, but theoretical results and explanations on crossover are still scarce.
There are some results for simple artificial test functions, where it was proven that a GA asymptotically outperforms an EA without crossover~\cite{JansenWegener:j:02,Jansen2005c,Watson2007,Storch2004,NeumannORS:c:11,KotzingST:c:11:crossover}
and the other way around~\cite{Richter2008}.
However, these artificial test functions are typically tailored to the specific algorithm and proof technique
and the results give little insight on the advantage of sexual reproduction on realistic problems occurring in artificial intelligence.
There are also a few theoretical results for problem-specific algorithms and representations,
namely coloring problems inspired by the Ising model~\cite{Sudholt2005} and the  
all-pairs shortest path problem~\cite{DoerrHK12}. 
For a nice overview of different aspects where populations and sex are beneficial for optimization of
static fitness functions, see~\cite{PrugelBennett:j:10}.

\medskip
\noindent\textbf{Noisy search.}
Heuristic optimization methods are typically not used for simple problems, but for rather difficult problems
in \emph{uncertain environments}. Evolutionary algorithms are very popular in settings including uncertainties;
see~\cite{Bia-Dor-Gam-Gut:j:09} for a survey on examples in combinatorial optimization, but also~\cite{JinBranke:j:05:robustSurvey} for an excellent survey also discussing different sources of uncertainty.
Uncertainty can be modeled by a \emph{probabilistic} fitness function, that is,
a search point can have different fitness values each time it is evaluated. 
One way to deal with this is to replace fitness evaluations with an average of a (large) sample of fitness evaluations and then proceed as if there was no noise. 
We take a different route, accept the noise, and try to analyze how much noise can be overcome by EAs without further modifications.
To do this in a rigorous manner, we assume \emph{additive posterior noise}, that is,
each time the fitness value of a search point is evaluated, we add a noise value drawn from some distribution.
This model was studied in evolutionary algorithms without crossover
in \cite{Gutjahr96,Sud-Thy:j:12,Doe-Hot-Koe:c:12,Fel-Koe:c:13}.

\medskip
\noindent\textbf{Our results on graceful scaling.} 
It has been observed that evolutionary algorithms benefit from sexual recombination on simple static problems.
It has also been observed that evolutionary algorithms (EAs) work in uncertain environments.
The important question, whether and how sexual recombination helps EAs on noisy problems,
remained open so far.  We introduce the concept of \emph{graceful scaling}~(Def.~\ref{def:graceful})
to measure how well a black-box optimization algorithm can handle noise.
We first prove a sufficient condition for when a noise model is intractable for optimization by a the classical \EA{\mu\!+\!1} (Theorem~\ref{thm:mutationTooMuchNoise}) and show that this implies that this simple
asexual algorithm does not scale gracefully for large Gaussian noise (Corollary~\ref{cor:lowerBoundGaussian}).
On the other hand, we study the compact GA (cGA), which strongly relies on recombination,
and prove that this sexual algorithm can handle noise gracefully (Theorem~\ref{thm:cga}).
These asymptotic results are complemented and matched by corresponding
experiments (Section~\ref{sec:experiments}).  We observe empirically that especially the
noise-oblivious variant of the cGA, which has no knowledge of properties of the added noise,
performs especially well. This confirms our theoretical finding that sexual recombination is
especially powerful in uncertain environments.

\onlyARXIV{
\medskip
\textbf{Biological motivation.}
Another motivation for our work comes from a biological perspective. The exact analysis of sexual recombination in both natural biological populations and in evolutionary computation is extremely difficult. In the field of population genetics, researchers often study the effects of recombination by describing the dynamics of natural selection on a freely recombining population under \emph{linkage equilibrium} in terms of the change in allele frequencies. Recently, several researchers have noticed a connection between these models and optimization algorithms such as EDAs~\cite{Muehlenbein1996recombination}  from the evolutionary computation community and the Multiplicative Weights Update Algorithm (MWUA)~\cite{AroraHK12}
also known from statistical machine learning~\cite{Chastain2014algorithms,Barton2014diverse}. The cGA is an EDA that tracks allele frequencies by simulating a population of $K$ individuals undergoing \emph{gene pool recombination}~\cite{Muehlenbein1996genepool} where offspring are produced essentially by performing crossover with all $K$ individuals as parents. In this way, the cGA is reasonably similar to models used in population genetics for studying sexual recombining populations, and thus we hope that our results can illuminate some of the utility of crossover in the presence of noisy signals for adaptation.
}


\section{Preliminaries}
Let $F$ be a family of pseudo-Boolean functions $(F_n)_{n \in \Na}$ where each $F_n$ is a set of functions $f\colon\{0,1\}^n \to \Re$. Let $D$ be a family of distributions $(D_v)_{v \in \Re}$ such that for all $D_v \in D$, $\E(D_v) = 0$. 
We define $F$ with additive posterior $D$-noise as the set $F[D] := \{f_n + D_v \colon f_n \in F_n, D_v \in D\}$.

\begin{definition}
\label{def:graceful}
  An algorithm $A$ \emph{scales gracefully with noise} on $F[D]$ if there is a polynomial $q$ such that, for all $g_{n,v} = f_n + D_v \in F[D]$, there exists a parameter setting $p$ such that $A(p)$ finds the optimum of $f_n$ using at most $q(n,v)$ calls to $g_{n,v}$.
\end{definition}

In the remainder of the paper, we will study a particular function class (\om) and a particular noise distribution (Gaussian, parametrized by the variance).
Let $\sigma^2 \geq 0$. We define the noisy \om function $\fnoise \colon \{0,1\}^n \to \Re := x \mapsto \|x\|_{1} + Z$ where $\|x\|_{1} := |\{i\colon x_i=1\}|$ and $Z$ is a normally distributed random variable  $Z \sim \mathcal{N}(0,\sigma^2)$ with zero mean and variance $\sigma^2$. 

The following proposition gives tail bounds for $Z$ by using standard estimates of the complementary error function~\cite{wolfram:erfc}.
\begin{proposition}
  \label{prp:tail}
Let $Z$ be a zero-mean Gaussian random variable with variance $\sigma^2$. For all $t > 0$ we have
\[
\Pr\left(Z < -t\right) = \frac{1}{2}\erfc\left(\frac{t}{\sigma\sqrt{2}}\right) \leq  \frac{1}{2}e^{-t^2/(2\sigma^2)}
\]
and asymptotically for large $t > 0$,
\[
\Pr\left(Z < -t\right) = \frac{1}{1+o(1)} \frac{\sigma}{\sqrt{2\pi}t}e^{-t^2/(2\sigma^2)}.
\]
\end{proposition}

\begin{definition}
    Let $x,y \in \{0,1\}^n$. Without loss of generality, suppose $\|x\|_1 - \|y\|_1 = \ell \geq 0$. Since $\fnoise$ is a function of unitation, the probability that it misclassifies $y$ as superior to $x$ depends only on the so-called phenotypic distance $\ell$. We define $\misclass\colon [n] \cup \{0\} \to [0,1]$ as
\[
\misclass(\ell) = \begin{cases}
  1/2 & \ell = 0,\\
  \Pr(\mathcal{E} \mid \|x\|_1 - \|y\|_1 = \ell) & \ell > 0;
\end{cases}
\]
where $\mathcal{E}$ is the event that $\fnoise(x) < \fnoise(y)$.
\end{definition}

\begin{lemma}
  \label{lem:error}  
  For any $0 \leq \ell < n$, $\misclass(\ell) > \misclass(\ell+1)$. Moreover, assuming $\sigma^2 > 0$, 
  \[
  \misclass(\ell) \leq \frac{1}{2}\left(1 - \Omega(\sigma^{-2})\right)
  \]
\end{lemma}
\begin{proof}
Let $x$ and $y$ be chosen arbitrarily from the set of all length-$n$ binary strings pairs with $\|x\|_1 - \|y\|_1 = \ell$ for any $\ell \in [n]$. The event that $\fnoise$ incorrectly classifies $y$ as superior to $x$ is equivalent to the event $\fnoise(x) < \fnoise(y)$. 
\[
\Pr(\fnoise(x) < \fnoise(y)) = \Pr\left(\ell + (Z_1 - Z_2) < 0\right),
\]
where $Z_1, Z_2 \sim \mathcal{N}(0,\sigma^2)$ are independent identically distributed. 
Letting $Z^* := Z_1 - Z_2$, we have $Z^* \sim \mathcal{N}(0,2\sigma^2)$ and $\misclass(\ell) = \Pr(Z^* < -\ell)$. Furthermore, $\misclass(\ell+1) = \Pr(Z^* < -(\ell + 1)) < \Pr(Z^* < -\ell) = \misclass(\ell)$. Finally, $\Pr\left(Z^* < -\ell \right) \leq (1/2)e^{-\ell^2/(4\sigma^2)} \leq (1/2)e^{-1/(4\sigma^2)}$ where we have applied Proposition~\ref{prp:tail}. The claim follows from the bound $1-x > e^{-x/(1-x)}$.
\end{proof}
A sequence of events $\{\mathcal{E}_n\}$ is said to hold \emph{with
  high probability} (\whp) if $\lim_{n \to \infty} \Pr(\mathcal{E}_n) = 1$.

\subsection{Algorithms}\label{sec:algorithms}
Algorithms that operate in the presence of noise often depend on \textit{a priori} knowledge of the noise intensity (measured by the variance). In such cases, the following scheme can always be used to transform such algorithms into one that has no knowledge of the noise character. Suppose $A(\sigma^2)$ is an algorithm that solves a noisy function with variance at most $\sigma^2$ within $T_\delta(\sigma^2)$ steps with probability at least $1 - \delta$. A \emph{noise-oblivious scheme} for $A$ is as follows.
\onlyARXIV{%
\begin{algorithm2e}
  \SetAlgoSkip{tinyskip}
  \SetKwFor{For}{repeat}{}{}
  $i \gets 0$\;
  \For{until solution found}{%
    Run $A(2^i)$ for $T_\delta(2^i)$ steps\;
    $i \gets i+1$\;
  }
\caption{Noise-oblivious scheme for $A$}
\label{alg:noise-oblivious}
\end{algorithm2e}
}
\onlyIJCAI{%
  Until a solution is found, run $A(2^i)$ for $T_{\delta}(2^i)$ steps for each $i \in (0,1,\ldots)$.
}
\onlyIJCAI{%
If an algorithm $A$ scales gracefully with noise, then the noise oblivious scheme for $A$ scales gracefully with noise. The following claim holds by a simple inductive argument.
}
\begin{claim}
  Suppose $f_{n,v} \in F[D]$ is a function with unknown variance $v$. Fixing $n$, assume $T_\delta$ grows at least linearly, but uniformly so. Then for any $s \in \Z^+$, the noise-oblivious scheme optimizes $f_{n,v}$ in at most $T_\delta(2^s v)$ steps with probability  at least $1 - \delta^s$. 
\end{claim}
\onlyARXIV{%
\begin{proof}
  By the assumptions on $T_\delta$, for all $c,x$, $cT_\delta(x) \leq T_\delta(cx)$ and so by induction, for any $k \in \Na$, $\sum_{i=0}^k T_\delta(2^i) \leq T_\delta(2^{k+1})$. Let phase $i$ be the $i$-th time in the for loop of Algorithm~\ref{alg:noise-oblivious}. We pessimistically suppose that the noise-oblivious scheme has not found a solution by phase $\log v - 1$. Then for the next $s$ phases, the proposed variance is at least $2^{\log v} = v$ and the probability that one of these phases is successful is at least $1 - \delta^s$. The total number of steps is at most $\sum_{i=0}^{\log v + s - 1} T_\delta(2^i) \leq T_\delta(2^s v)$.
\end{proof}
}

The \EA{\mu+1}, defined in Algorithm~\ref{alg:ea}, is a simple mutation-only evolutionary algorithm that maintains a population of $\mu$ solutions and uses elitist survival selection.
\begin{algorithm2e}
  $t \gets 1$\;
  $P_t \gets \mu$ elements of $\{0,1\}^n$ u.a.r.\;
  \While{termination criterion not met}{
    Select $x \in P_t$ u.a.r.\;
    Create $y$ by flipping each bit of $x$ w/ probability $1/n$\;
    $P_{t+1} \gets P_t \cup \{y\} \setminus \{z\}$ where $f(z) \leq f(v) \forall v \in P_t$\;
    $t \gets t+1$\;
  }
\caption{\label{alg:ea} The \EA{\mu+1}}
\end{algorithm2e}

The compact genetic algorithm (cGA)~\cite{Harik1999cga} is a genetic algorithm that maintains a population of size $K$ \emph{implicitly} in memory. Rather than storing each individual separately, the cGA only keeps track of population \emph{allele frequencies} and updates these frequencies during evolution. Offspring are generated according to these allele frequencies, which is similar to what occurs in a sexually-recombining population. Indeed, the offspring generation procedure can be viewed as so-called \emph{gene pool recombination} introduced by M{\"u}hlenbein and Paa{\ss}~\shortcite{Muehlenbein1996genepool} in which all $K$ members participate in recombination. Since the cGA evolves a probability distribution, it is also a type of \emph{estimation of distribution algorithm} (EDA). The correspondence between EDAs and models of sexually recombining populations has already been noted~\cite{Muehlenbein1996recombination}, and Harik et al.~\shortcite{Harik1999cga} demonstrate empirically that the behavior of the cGA is equivalent to a simple genetic algorithm at least on simple problems.

The first rigorous analysis of the cGA is due to Droste~\shortcite{Droste2006cga} who gave a general runtime lower bound for all pseudo-Boolean functions, and a general upper bound for all linear pseudo-Boolean functions. Defined in Algorithm~\ref{alg:cga}, the cGA maintains for all times $t \in \Na$ a frequency vector $(p_{1,t},p_{2,t},\ldots,p_{n,t}) \in [0,1]^n$. In the $t$-th iteration, two strings $x$ and $y$ are sampled independently from this distribution where $\Pr(x = z) = \Pr(y = z) = \left(\prod_{i \colon z_i = 1} p_{i,t}\right)\times\left(\prod_{i \colon z_i = 0} (1 - p_{i,t})\right)$
for all $z \in \{0,1\}^n$. The cGA then compares the objective values of $x$ and $y$, and updates the distribution by advancing $p_{i,t}$ toward the component of the winning string by an additive term.
\begin{algorithm2e}
  $t \gets 1$\;
  $p_{1,t} \gets p_{2,t} \gets \cdots \gets p_{n,t} \gets 1/2$\;
  \While{termination criterion not met}{
    \For{$i \in \{1,\ldots,n\}$\label{li:x}}{
      $x_i \gets 1$ w/ prob.\ $p_{i,t}$, $x_i \gets 0$ w/ prob.\ $1 - p_{i,t}$
    }
  \For{$i \in \{1,\ldots,n\}$\label{li:y}}{
    $y_i \gets 1$ w/ prob.\ $p_{i,t}$, $y_i \gets 0$ w/ prob.\ $1 - p_{i,t}$
  }
  \lIf{$f(x) < f(y)$\label{li:eval}}{swap $x$ and $y$}
  \For{$i \in \{1,\ldots,n\}$}{
    \lIf{$x_i > y_i$}{$p_{i,t+1} \gets p_{i,t} + 1/K$\;}
    \lIf{$x_i < y_i$}{$p_{i,t+1} \gets p_{i,t} - 1/K$\;}
    \lIf{$x_i = y_i$}{$p_{i,t+1} \gets p_{i,t}$\;}
  }
  $t \gets t+1$\;
}
\caption{\label{alg:cga} The compact GA}
\end{algorithm2e}


\section{Results}
\label{sec:Results}
We derive rigorous bounds on the optimization time, defined as of the first hitting time of the process to the true optimal solution ($1^n$) of $\fnoise$, on a mutation-only based approach and the compact genetic algorithm. 


\subsection{Mutation-based Approach}
In this section we consider the \EA{\mu+1}. We will first, in Theorem~\ref{thm:mutationTooMuchNoise}, give a sufficient condition for when a noise model is intractable for optimization by a \EA{\mu+1}. Then we will show that, in the case of additive posterior noise sampled from a Gaussian distribution, this condition is fulfilled if the noise is large enough, showing that the \EA{\mu+1} cannot deal with arbitrary Gaussian noise (see Corollary~\ref{cor:lowerBoundGaussian}).

\begin{theorem}
\label{thm:mutationTooMuchNoise}
Let $\mu \geq 1$ and $D$ a distribution on $\Re$. Let $Y$ be the random variable describing the minimum over $\mu$ independent copies of $D$. Suppose
$$
\Pr(Y > D+n) \geq \frac{1}{2(\mu+1)}.
$$
Consider optimization of \om with reevaluated additive posterior noise from $D$ by \EA{\mu+1} without crossover. Then, for $\mu$ bounded from above by a polynomial, the optimum will \emph{not} be evaluated after polynomially many iterations \whp
\end{theorem}
\begin{proof} 
For all $t$ and all $i \leq n$ let $X^t_i$ be the random variable describing the \emph{proportion of individuals} in the population of iteration $t$ with exactly $i$ $1$s. Let $c = 800$, $b = 20$, $a = (c-1)/c$ and $a' = (c-2)/c$. We show by induction on $t$ that
$$
\forall t, \forall i \geq a n\colon E[X^t_i] \leq b^{a n-i}.
$$
In other words, the expected number of individuals with $i$ $1$s is decaying exponentially with $i$ after $an$. This will give the desired result with a simple union bound over polynomially many time steps. \onlyIJCAI{In the interest of saving space, we only skip the remainder of the proof.}
\onlyARXIV{

The claim holds at the start of the algorithm with an application of Hoeffding's Inequality for the number of $1$s in a random individual. Fix some value $t$ and suppose the claim holds for that $t$. Let some value $i \geq an$ be given and let $x = b^{a n-i}$. We will now show $E[X^{t+1}_i] \leq x$ by considering one generation of the \EA{\mu+1} without crossover.

We distinguish four cases depending on whether an individual with less than $a' n$ $1$s has been selected for reproduction, with $i-k$ $1$s for some $k$ with $1 \leq k \leq n/c$, with exactly $i$ $1$s or with strictly more than $i$ $1$s. For each of these cases we estimate the number of individuals that can be chosen to reproduce, as well as the probability for such an individual to produce an offspring with exactly $i$ $1$s. The following table gives upper bounds for both values in all four cases; we will justify all these values below.

\begin{center}
\begin{tabular}{ccc}
\noalign{\hrule height .3mm}
 & Proportion  & Probability\vspace*{.04cm}\\
\noalign{\hrule height .3mm}
$<a' n$ & $1$ & $2^{-\Omega(n\ln n)}$\\
$=i-k$ & $xb^k$ & $(2/c)^{k}$\\
$=i$ & $x$ & $1/e + 1/(c-1)$\\
$>i$ & $x/(b-1)$ & $1$\vspace*{.04cm}\\
\noalign{\hrule height .3mm}
\end{tabular}
\end{center}

Clearly the proportion of individuals with $<a' n$ $1$s is bounded from above by $1$; for such an individual with $m$ $0$s, at least half of these $0$s need to flip, which has a probability of at most $2^m/n^{m/2} = 2^{-\Omega(n\ln n)}$, using $m \geq n/c$. For any $k < n/c$, we get a bound of $xb^k$ for the number of individuals with exactly $i-k$ $1$s from the induction hypothesis; as these individuals have at most $2n/c$ many $0$s, the probability of flipping at least $k$ of these to $1$ is $\leq (2/c)^{k}$. For an individual with exactly $i$ $1$s to create an offspring with exactly $i$ $1$s, we can either not flip any bit (with a probability tending to $1/e$) or we flip as many $1$s as $0$s; flipping $k$ $1$s has a probability of at most $1/c$ (as $i \geq an$), thus we can bound the probability of creating an offspring with exactly $i$ $1$s by
$$
1/e + \sum_{k=1}^{\infty} c^{-k} = 1/e + 1/(c-1).
$$
With a similar geometric sum we get that the number of individuals with $>i$ $1$s is, using the induction hypothesis, at most $x/(b-1)$.

From the table we can now deduce that the probability of producing an offspring with exactly $i$ $1$s in iteration $t$ is at most
\begin{equation*}\label{eq:BoundNumberOne}
2^{-\Oh(n\ln n)} + x\left(\frac{1}{e} + \frac{1}{c-1} + \frac{1}{b-1} + \sum_{k=1}^{\infty}\left(\frac{2b}{c}\right)^k \right)
\end{equation*}
Using $x \geq b^{-n/c}$ we see that $2^{-\Oh(n\ln n)}$ has asymptotically no impact on the sum. Furthermore, from our choice of $b$ and $c$, we have
$$
\frac{1}{e} + \frac{1}{c-1} + \frac{1}{b-1} + \frac{1}{c/(2b)-1} < 1/2.
$$
Thus, we have that we get less than $x/2$ individuals with exactly $i$ $1$s in expectation, while the premise of the theorem gives that any individual has a probability of at least $1/2$ to die in any given iteration. This shows that $E[X^{t+1}_i]$ cannot go above $x$.}
\end{proof}

We apply Theorem~\ref{thm:mutationTooMuchNoise} to show that large noise levels make it impossible for the \EA{\mu+1} to efficiently optimize.
\begin{corollary}\label{cor:lowerBoundGaussian}
Consider optimization of 
\fnoise
by \EA{\mu+1} without crossover. Suppose $\sigma^2 \geq n^3$ and $\mu$ bounded from above by a polynomial in $n$. Then the optimum will \emph{not} be evaluated after polynomially many iterations \whp
\end{corollary}
\onlyARXIV{\begin{proof}
We set up to use Theorem~\ref{thm:mutationTooMuchNoise}. Let $D \sim \mathcal{N}(0,\sigma^2)$ and let $Y$ be the minimum over $\mu$ independent copies of $D$. We want to bound $\Pr(Y > D+n)$. To that end we let $t_1 < 0 $ and $t_1 < t_0$ be such that $\Pr(D <  t_0) = 0.75/\mu$ and $\Pr(D <  t_1) = 1.5/\mu$. Let $A$ be the event that $Y > t_1$ and $t_0 -n< D < t_1 - n$ and let $B$ be the event that $Y > t_0$ and $D < t_0 - n$. Clearly, the events $A$ and $B$ are disjoint and are contained in the event that $Y > D+n$. From the asymptotic bounds stated in Proposition~\ref{prp:tail} and the lower bound on $\sigma^2$ we see that $t_0 - n \geq t_0(1+n^{-0.5})$; similarly, $t_0(1+n^{-0.5}) \leq t_1 - n \leq t_0(1-n^{-0.5})$.
This gives that $\Pr(D <  t_0-n)$ and $\Pr(t_0 -n< D < t_1 - n)$ are both asymptotically $0.75/\mu$, as they would be without the ``$-n$''-terms; this uses the bound on $\mu$.
Thus, we have asymptotically
\begin{align*}
 \Pr(Y > D+n)
 & \geq  \Pr(A) + \Pr(B)\\
 & \geq  \frac{0.75}{\mu}\left(1- \frac{1.5}{\mu} \right)^\mu + \frac{0.75}{\mu}\left(1- \frac{0.75}{\mu} \right)^\mu\\
 & \geq  \frac{1}{2(\mu+1)}.
\end{align*}
The last step uses the bound $1-x > \exp(-x/(1-x))$.
\end{proof}}


\subsection{Compact GA}
\label{sec:cga}
Let $T^\star$ be the optimization time of the cGA on $\fnoise$, namely, the first time that it generates the underlying ``true'' optimal solution $1^n$. We consider the stochastic process $X_t = n - \sum_{i=1}^n p_{i,t}$ and bound the optimization time by $T = \inf\{t > 0 \colon X_t = 0\}$. Clearly $T^\star \leq T$ since the cGA produces $1^n$ in the $T$-th iteration almost surely.
However, $T^\star$ and $T$ can be infinite when there is a $t < T^\star$ where $p_{i,t}=0$ since the process can never subsequently generate any string $x$ with $x_i = 1$.
To circumvent this, Droste~\shortcite{Droste2006cga} estimates $\E(T^\star)$ conditioned on the event that $T^\star$ is finite, and then bounds the probability of finite $T^\star$. In this paper, we will prove that as long as $K$ is large enough, the optimization time is finite (indeed, polynomial) with high probability. 

\onlyARXIV{%
The following lemma is due to von Bahr and Esseen~\shortcite{vonBahr1965inequalities} and states an exact equality for the first absolute moment of a random variable $Z$ in terms of its characteristic function $\varphi_Z(t) = \E(e^{itZ})$.
}
\onlyIJCAI{%
  The \emph{characteristic function} of a random variable $Z$ is defined as $\varphi_Z(t) = \E(e^{itZ})$. We need the following result.
}
\begin{lemma}[special case of Lemma 2 of \cite{vonBahr1965inequalities}]
  \label{lem:vonBahr}
  Let $Z$ be a random variable with $\E(|Z|) < \infty$. Then
  \[
  \E(|Z|) = \frac{1}{\pi} \int_{-\infty}^{\infty}
  \left(1 - \mathfrak{R}(\varphi_Z(t))\right)/t^2\,\mathrm{d}t,
  \]
  where $\mathfrak{R}(z)$ is the real part of $z \in \Co$.
\end{lemma}

\begin{lemma}
  \label{lem:Z-properties}
  Let $0 < a < 1$ be a constant. Consider a random variable $Z = Z_1 + Z_2 + \cdots + Z_n$, each $Z_i$ independent,
  \[
  Z_i = \begin{cases}
    1 & \text{with probability $p_i(1 - p_i)$,}\\
    -1 & \text{with probability $p_i(1 - p_i)$,}\\
    0 & \text{with probability $1 - 2p_i(1 - p_i)$;}
  \end{cases}
  \]
  with $a \leq p_i \leq 1$ for every $i \in \{1,\ldots,n\}$. Then
  $\Pr(Z = 0) \geq 1/(4\sqrt{n})$, and
  \[
  \E(|Z|) \geq a\sqrt{2/n}\left(n - \sum_{i=1}^n p_i\right).
  \]
\end{lemma}
\begin{proof}
  \onlyARXIV{%
    Let $\xi = |Z_1| + |Z_2| + \cdots + |Z_n|$. Then $\xi$ is distributed as a Poisson-Binomial distribution with each success probability equal to $2p_i(1-p_i)$ and
    \[
    \Pr(Z = 0) = \sum_{k = 0}^{n} \Pr(\xi = k) \binom{k}{k/2} 2^{-k}
    \]
    where $\binom{k}{k/2} = 0$ if $k \equiv 1 \pmod{2}$. This is the joint probability that exactly $k$ of the $Z_i$ variables are nonzero, and exactly half of these are selected to be negative, the other half positive. Since $\binom{k}{k/2}$ vanishes at odd $i$, we can write
    \[
    \Pr(Z = 0) = \sum_{k=0}^{\lfloor n/2 \rfloor} \Pr(\xi = 2k) \binom{2k}{k} 2^{-2k}.
    \]
    $\binom{2k}{k}$ is the $k$-th central binomial coefficient, for which we have the well-known bound $\frac{2^{2k}}{\sqrt{4k}} \leq \binom{2k}{k}$, so we can write
\begin{equation}
  \Pr(Z = 0) \geq \Pr(\xi = 0) + \sum_{k=1}^{\lfloor n/2 \rfloor} \Pr(\xi = 2k)\frac{1}{2\sqrt{k}} 
\geq \frac{1}{2\sqrt{n}}\Pr(\xi\text{~is even}),  \label{eq:bounds}
\end{equation}
since $\frac{1}{2\sqrt{n}} \leq \frac{1}{2\sqrt{k}} \leq 1$.
To finish the proof, note that for any integer random variable $X$,
$\Pr(X\text{~is even}) = (1 + G(-1))/2$,
where $G(z) = \E(z^X)$ is the probability generating function for $X$. For a Poisson-Binomial distribution with success probabilities $q_1, q_2, \ldots, q_n$, 
$G(z) = \prod_{i=1}^n (1 - q_i + q_i z)$, so, 
\[
\Pr(\xi\text{~is even}) = \frac{1}{2}\left(1 + \prod_{i=1}^n (1 - 2q_i)\right).
\]
Finally, since $q_i = 2p_i(1-p_i) \leq 1/2$ for all $i \in \{1,\ldots,n\}$, $\Pr(\xi\text{~is even}) \geq 1/2$ and the claimed bound on $\Pr(Z = 0)$ follows from~\eqref{eq:bounds}.
}
\onlyIJCAI{%
  Let $\xi = |Z_1| + |Z_2| + \cdots + |Z_n|$. Then $\xi$ is distributed as a Poisson-Binomial distribution with each success probability equal to $2p_i(1-p_i)$. Hence,
\begin{align*}
    \Pr(Z = 0) &= \sum_{k=0}^{\lfloor n/2 \rfloor} \Pr(\xi = 2k) \binom{2k}{k} 2^{-2k}\\
    &\geq \Pr(\xi = 0) + \sum_{k=1}^{\lfloor n/2 \rfloor} \Pr(\xi = 2k)\frac{1}{2\sqrt{k}}\\
    &\geq \frac{1}{2\sqrt{n}}\Pr(\xi\text{~is even}) \geq 1/(4\sqrt{n}).
\end{align*}
The first inequality follows from well-known bounds on the central binomial coefficient.
The final inequality uses the probability generating function of the Poisson-Binomial distribution, namely,
$\Pr(\xi\text{~is even}) = \frac{1}{2}\left(1 + \prod_{i=1}^n (1 - 2q_i)\right) \geq 1/2$ since $q_i = 2p_i(1-p_i) \leq 1/2$ for each $i \in \{1,\ldots,n\}$.
}

We now bound the first absolute moment of $Z$ from below.
For every $S \subseteq [n]$, denote as $\mathcal{E}_S$ the event that $|Z_i| = 1 \iff i \in S$. We first calculate the expectation of $|Z|$ conditioned on these events. Since the probabilities $p_i$ are independent,
  \begin{align*}
  \E(e^{itZ} \mid \mathcal{E}_S) &= \prod_{j=1}^n \E(e^{itZ_j} \mid \mathcal{E}_S)\\
    \E(e^{itZ} \mid \mathcal{E}_S) &= \prod_{j=1}^n \left([j \in S]\left(\frac{e^{it}}{2} + \frac{e^{-it}}{2}\right) + [j \not\in S]\right)\\ 
  &= \prod_{j \in S} \cos t = (\cos t)^{|S|}
  \end{align*}
  where $[P]$ is the Iverson bracket. So by Lemma~\ref{lem:vonBahr},
\[
 \E(|Z| \mid \mathcal{E}_S) = \frac{1}{\pi} \int_{-\infty}^\infty \frac{1 - (\cos t)^{|S|}}{t^2}\, \mathrm{d}t = g(|S|),
\]
where $g(k) = 2\lceil k/2 \rceil \binom{2\lceil k/2 \rceil}{\lceil k/2 \rceil} 2^{-2 \lceil k/2 \rceil}$. Again applying bounds on the central binomial coefficient,
$g(k) \geq \sqrt{\lceil k/2 \rceil} \geq \sqrt{k/2}$. 
By the law of total expectation,
\onlyARXIV{%
\begin{equation}
  \label{eq:expectation-Z}
  \E(|Z|) = 
  \sum_{k=1}^n g(k) \sum_{S \subseteq [n] \colon |S| = k} \Pr(\mathcal{E}_S)
  \geq \frac{1}{\sqrt{2n}} \sum_{k=1}^n k \sum_{S \subseteq [n] \colon |S| = k} \Pr(\mathcal{E}_S)
  = \E(\xi)/\sqrt{2n},
\end{equation}
}
\onlyIJCAI{%
  \begin{multline}
    \label{eq:expectation-Z}
  \E(|Z|) = 
  \sum_{k=1}^n g(k) \sum_{S \subseteq [n] \colon |S| = k} \Pr(\mathcal{E}_S)\\
  \geq \frac{1}{\sqrt{2n}} \sum_{k=1}^n k \sum_{S \subseteq [n] \colon |S| = k} \Pr(\mathcal{E}_S) = \frac{\E(\xi)}{\sqrt{2n}},
\end{multline}
}
Since $\xi$ follows a Poisson-Binomial distribution with the $i$-th success probability equal to $2p_i(1-p_i)$,
and every $p_i \geq a$,
\[
\E(\xi) = \sum_{i=1}^n 2p_i(1 - p_i) \geq 2a\left(n - \sum_{i=1}^n p_i\right).
\]
Substituting this inequality into~\eqref{eq:expectation-Z} completes the proof.
\end{proof}

\begin{lemma}
\label{lem:drift}
Consider the cGA optimizing $\fnoise$ and let $X_t$ be the stochastic process defined above. Assume that there exists a constant $a > 0$ such that $p_{i,t} \geq a$ for all $i \in \{1,\ldots,n\}$ and that $X_t > 0$, then $\E\left(X_{t} - X_{t+1} \mid X_t \right) \geq \delta X_t$ where $1/\delta = O{\left(\sigma^2K\sqrt{n}\right)}$.
\end{lemma}

\begin{proof}
Let $x$ and $y$ be the offspring generated in iteration $t$ and $Z_t = \|x\|_1 - \|y\|_1$. Then $Z_t = Z_{1,t} + \cdots + Z_{n,t}$ where \[
Z_{i,t} = 
\begin{cases}
  -1 & \text{if $x_i = 0$ and $y_i = 1$,}\\
  0 & \text{if $x_i = y_i$,}\\
  1 & \text{if $x_i =1$ and $y_i = 0$;}
\end{cases}
\]
Let $\mathcal{E}$ denote the event that in line~\ref{li:eval}, the evaluation of $\fnoise$ correctly ranks $x$ and $y$. Without loss of generality, suppose $\|x\|_1 \geq \|y\|_1$. 
Then $\E(X_{t+1} - X_t \mid X_t, \mathcal{E}) = \E(|Z_t|)/K$.
On the other hand, if $\fnoise(x)$ evaluates to at most $\fnoise(y)$ during iteration $t$, the roles above are swapped and $\E(X_{t+1} - X_t \mid X_t \land \overline{\mathcal{E}}) = -\E(|Z_t|)/K$.
  By the law of total expectation, 
\begin{equation}
\label{eq:drift}
\E(X_{t} - X_{t+1} \mid X_t) = \frac{\E(|Z_t|)}{K}\left(1 - 2\Pr(\overline{\mathcal{E}})\right).
\end{equation}
For any $i \in [n]$, $\Pr(Z_{i,t} = 1) = \Pr(Z_{i,t} = -1) = p_{i,t}(1 - p_{i,t})$ and $\Pr(Z_{i,t} = 0)$ is the inverse. Since we have assumed each $p_{i,t} \geq a$, we can apply Lemma~\ref{lem:Z-properties} to obtain
  \begin{equation}
    \label{eq:Z-bound}
  \E(|Z_t|) \geq a\sqrt{2/n}\left( n - \sum_{i=1}^n p_{i,t}\right)\\
  = a X_t\sqrt{2/n}.
  \end{equation}
  To complete the proof, we substitute the inequality in Equation~\eqref{eq:Z-bound} into Equation~\eqref{eq:drift} and use Lemma~\ref{lem:error} to bound $\Pr(\overline{\mathcal{E}}) = \misclass(|\|x\|_1 - \|y\|_1|)$ from above. 
\end{proof}

\begin{lemma}
  \label{lem:extinction}
  Consider the cGA optimizing $\fnoise$ with $\sigma^2 > 0$. Let $0 < a < 1/2$ be an arbitrary constant and $T' = \min\{ t \geq 0 \colon \exists i \in [n], p_{i,t} \leq a \}$. If $K = \omega(\sigma^2 \sqrt{n} \log n)$, then for every polynomial $\poly(n)$, $n$ sufficiently large, $\Pr(T' < \poly(n))$ is superpolynomially small.
\end{lemma}
\begin{proof}
  Let $i \in [n]$ be arbitrary. Let $\{Y_t \colon t > 0\}$ be the stochastic process $Y_t = \left(1/2 - p_{i,t}\right)K$.
  We first argue that
  \begin{equation}
    \label{eq:claim}
    \E(Y_t \mid Y_1, \ldots, Y_{t-1}) \leq Y_{t-1} - \Omega(\sigma^{-2})\frac{\Pr(x_i \neq y_i)}{\sqrt{n}}.
    \tag{$\star$}
  \end{equation}
  Let $x$ and $y$ be the strings generated in iteration $t$ of the cGA (lines~\ref{li:x} and~\ref{li:y} of Algorithm~\ref{alg:cga}). We define $\hat{x} = (x_1, x_2, \ldots, x_{i-1}, x_{i+1}, \ldots x_n)$ to be the substring of $x$ constructed by removing the $i$-th element and $\hat{y}$ similarly. Since each element of $x$ and $y$ is constructed independently, we can regard $\hat{x}$, $\hat{y}$, $x_i$, and $y_i$ to be independent.

Note that $\E(Y_t \mid Y_1, \ldots, Y_{t-1}) = Y_{t-1} + \delta_t$ where $\delta_t \in \{-1,0,1\}$. Define $\hat{\ell} = \|\hat{x}\|_1 - \|\hat{y}\|_1$. We distinguish between the two events that $|\hat{\ell}|$ is nonzero or zero. 

\begin{description}[leftmargin=0pt,font=\normalfont\itshape]
\item [Case $|\hat{\ell}|> 0$.] Suppose without loss of generality that $\hat{\ell} > 0$ (i.e., $\|\hat{x}\|_1 > \|\hat{y}\|_1$). So, $\delta_t = 0$ if and only if $x_i = y_i$. Moreover, $\delta_t = -1$ only in the event that (a) $x_i = 1$ and $y_i = 0$ and $x$ is accepted (in which case $\ell = \hat{\ell} + 1$), or (b) $x_i = 0$ and $y_i = 1$ and $x$ is \emph{not} accepted (in which case $\ell = \hat{\ell} - 1$). Event (a) occurs only if $\fnoise$ does not misclassify $x$ and $y$, whereas event (b) occurs only if $\fnoise$ does misclassify $x$ and $y$. Thus, 
$\Pr(\delta_t = - 1) = \Pr(x_i =1,  y_i=0)\left(1 - \misclass(\hat{\ell}+1)\right) + \Pr(x_i =0, y_i=1)\misclass(\hat{\ell}-1)$.

Similarly, $\delta_t = 1$ only in the event that (a) $x_i = 1$ and $y_i = 0$ but $x$ is \emph{not} accepted because $x$ and $y$ were misclassified by $\fnoise$, or (b) $x_i = 0$ and $y_i = 1$ and $x$ is accepted because $\fnoise$ ranked $x$ and $y$ correctly. Thus,
$\Pr(\delta_t = 1) = \Pr(x_i = 1, y_i = 0)\misclass(\hat{\ell}+1)
 + \Pr(x_i = 0, y_i = 1)(1 - \misclass(\hat{\ell} - 1))$. 
Since $\Pr(x_i = 1, y_i = 0) = \Pr(x_i = 0, y_i = 1) = \Pr(x_i \neq y_i)/2$,
\[
  \E(\delta_t) = Pr(\delta_t = 1) - \Pr(\delta_t = -1)
  = -\Pr(x_i \neq y_i)\left(\misclass(\hat{\ell}-1) - \misclass(\hat{\ell} + 1)\right) < 0,
\]
where we apply Lemma~\ref{lem:error}. We conclude that in this case,
\[
\E(Y_t \mid \hat{\ell} \neq 0, Y_1, \ldots, Y_{t-1}) = Y_{t-1} + \E(\delta_t) < Y_{t-1}.
\]

\item [Case $\hat{\ell}  = 0$.] In this case, if $x_i = y_i$, then $x = y$ and there is zero drift. Otherwise, 
$x_i > y_i$ and so $\|x\|_1 - \|y\|_1 = 1$, or $y_i > x_i$ and $\|y\|_1 - \|x\|_1 = 1$. The drift in this case only depends on whether or not $\fnoise$ misclassifies $x$ and $y$. In particular,
$\Pr(\delta_t = -1) = \Pr(x_i = 1, y_i = 0)(1 - \misclass(1)) 
 + \Pr(x_i = 0, y_i = 1)(1 - \misclass(1))$, and
$\Pr(\delta_t = 1) = \Pr(x_i = 1, y_i = 0)\misclass(1) 
+ \Pr(x_1 = 0, y_i = 1)\misclass(1)$. By Lemma~\ref{lem:error},
\begin{align*}
\E(\delta_t) &= \Pr(\delta_t = 1) - \Pr(\delta_t = -1) \\
&= -\Pr(x_i \neq y_i)(1 - 2\misclass(1)) \\
&\leq
-\Omega(\sigma^{-2})\Pr(x_i \neq y_i).
\end{align*}
For this case, $\E(Y_t \mid \hat{\ell} = 0, Y_1, \ldots, Y_{t-1}) = Y_{t-1} + \E(\delta_t) \leq Y_{t-1} - \Omega(\sigma^{-2})\Pr(x_i \neq y_i)$.
\end{description}
Applying the law of total expectation, $\E(Y_t \mid Y_1, \ldots, Y_{t-1})$ is bounded above by
\[
Y_t - \Omega(\sigma^{-2})\Pr(x_i \neq y_i)\Pr(\hat{\ell} = 0) .
\]

It remains to bound $\Pr(\hat{\ell} = 0) = \Pr(\|\hat{x}\|_1 = \|\hat{y}\|_1)$. We define a random variable 
$Z = Z_2 + \cdots + Z_n$ where
 \[
 Z_j = \begin{cases}
   +1 & \text{if $x_j > y_j$,}\\
   0  & \text{if $x_j = y_j$,}\\
   -1 & \text{if $x_j < y_j$.}
 \end{cases}
 \]
 So $\Pr(\|\hat{x}\|_1 = \|\hat{y}\|_1) = \Pr(Z = 0) \geq 1/(4\sqrt{n-1})$ by Lemma~\ref{lem:Z-properties} since $0 \leq \Pr(x_j > y_j) = \Pr(x_j < y_j) = p_j(1 - p_j) \leq 1/2$ for all $j \in \{2,\ldots,n\}$, proving the claim in~\eqref{eq:claim}.

Note that $\{Y_t \colon t \in \Na\}$ is a Markov chain on $\{-K/2,-K/2+1,\ldots,K/2-1,K/2\}$ with $Y_1 = 0$. Let $T = \min\{t \colon Y_t > (1/2 - a)K\}$. In any iteration, if $x_i = y_i$, then $Y_t = Y_{t-1}$. Thus, for an estimate of the upper bounds of $T$, we can ignore self-loops in the chain.

More formally, let $\{\hat{Y}_t \colon t \in \Na\}$ be the restriction of $Y_t$ to iterations such that $Y_t \neq Y_{t-1}$. Similarly, let $\hat{T} = \min\{t \colon \hat{Y}_t > (1/2 - a)K\}$. The random variable $T$ stochastically dominates the random variable $\hat{T}$ since removing equal moves can only make the process hit faster, i.e., $\forall t \in \Na$, $\Pr(T > t) \geq \Pr(\hat{T} > t)$. Due to the above arguments,
\onlyARXIV{%
\begin{align*}
\E(\hat{Y}_t \mid \hat{Y}_1, \ldots, \hat{Y}_{t-1}) 
&= \E(\hat{Y}_t \mid x_i \neq y_i, \hat{Y}_1, \ldots, \hat{Y}_{t-1})\\
& = \hat{Y}_t - E(\delta_t \mid x_i \neq y_i)\\
&\leq \hat{Y}_t - \Omega(\sigma^{-2}/\sqrt{n}.
\end{align*}
}
\onlyIJCAI{%
\begin{align*}
\E(\hat{Y}_t \mid \hat{Y}_1, \ldots, \hat{Y}_{t-1}) 
 &= \hat{Y}_t - E(\delta_t \mid x_i \neq y_i)\\
&\leq \hat{Y}_t - \Omega(\sigma^{-2}/\sqrt{n}).
\end{align*}
}
By a refinement to the negative drift theorem of Oliveto and Witt~\shortcite{Oliveto2011simplified,Oliveto2012erratum} (cf Theorem 3 of \cite{Koetzing2014concentration}), since $Y_1 = \hat{Y}_1 = 0$ and $|\hat{Y}_t - \hat{Y}_{t+1}| = 1 < \sqrt{2}$, for all $s \geq 0$,
\[
\Pr(T \leq s) \leq \Pr(\hat{T} \leq s) \leq s \exp\left(-\frac{(1/2 - a)K|\epsilon|}{32}\right),
\]
with $\epsilon = -\Omega(\sigma^{-2}/\sqrt{n})$. Since $K = \omega(\sigma^2\sqrt{n} \log n)$, $\Pr(T \leq s) = s n^{-\omega(1)}$.

So, for any polynomial $s = \poly(n)$, with probability superpolynomially close to one, $Y_s$ has not yet reached a state larger than $(1/2 - a)K$, and so $p_{i,t} > a$ for all $1 \leq t \leq s$. As this holds for arbitrary $i$, applying a union bound retains a superpolynomially small probability that any of the $n$ frequencies have gone below $a$ by $s = \poly(n)$ steps.
\end{proof}

\begin{theorem}
  \label{thm:cga}
  Consider the cGA optimizing $\fnoise$ with variance $\sigma^2 > 0$ for any constant $c \geq 0$. If $K = \omega(\sigma^2 \sqrt{n} \log n)$, then with probability $1 - o(1)$, the cGA finds the optimum after $\Oh(K\sigma^2\sqrt{n} \log Kn)$ steps.
\end{theorem}
\begin{proof}
  
  We will consider the drift of the stochastic process $\{X_t \colon t \in \Na\}$ over the state space $S \subseteq \{0\} \cup [x_{\min}, x_{\max}]$ where $X_t = n - \sum_{i=1}^n p_{i,t}$. Hence, $x_{\min} = 1/K$.

Fix a constant $0 < a < 1/2$. We say the process has \emph{failed} by time $t$ if there exists some $s \leq t$ and some $i \in [n]$ such that $p_{i,s} \leq a$. Let $T = \min\{t \in \Na \colon X_t = 0\}$. Assuming the process never fails, by Lemma~\ref{lem:drift}, the drift of $\{X_t \colon t \in \Na\}$ in each step is bounded by $\E\left(X_{t} - X_{t+1} \mid X_t = s\right) \geq \delta X_t$ where $1/\delta = O{\left(\sigma^2K\sqrt{n}\right)}$. Hence, by tail bounds for the multiplicative drift theorem (see Doerr and Goldberg~\shortcite{Doerr2013adaptive}),
$\Pr\left(T \geq \left(\ln(X_1/x_{\min}) + r\right)/\delta\right) \leq e^{-r}$.
Choosing $r = d\ln n$ for any constant $d > 0$, the probability that $T = \Omega(K\sigma^2\sqrt{n} \log Kn)$ is at most $n^{-d}$.

Letting $\mathcal{E}$ be the event that the process has not failed by $\Oh(K\sigma^2\sqrt{n}\log Kn)$ steps, by the law of total probability, the hitting time of $X_t = 0$ is bounded by $\Oh(K\sigma^2\sqrt{n} \log Kn)$ with probability $(1 - n^{-d})\Pr(\mathcal{E}) = 1 - o(1)$ where we can apply Lemma~\ref{lem:extinction} to bound the probability of $\mathcal{E}$. 
\end{proof}


\section{Experiments}
\label{sec:experiments}

\definecolor{color1}{RGB}{218,124,48}
\definecolor{color2}{RGB}{204,37,41}
\definecolor{color3}{RGB}{0,0,0}
\definecolor{color4}{RGB}{57,106,177}

In Section~\ref{sec:Results} we proved that the cGA scales gracefully with noise (see Def.~\ref{def:graceful}) on a simple noisy pseudo-Boolean function, whereas a mutation-only EA fails when the noise variance is too high. In this section, we seek to compare the performance of the cGA with a baseline hillclimber that uses explicit resampling to reduce the noise variance. 

Our baseline hillclimber is called resampling randomized local search (reRLS). For a particular variance $\sigma^2$, reRLS estimates the true objective function value by performing $\Oh(\sigma^2 \log n)$ function calls for each search point~\cite{Teytaud2015noisy}. It then hillclimbs on the estimated true objective function by flipping a single bit in each iteration and accepting points with equal or better estimated objectives. Both reRLS and the cGA require knowledge of the true noise variance to collect enough samples (reRLS) or to set $K$ properly (cGA). We also investigate the performance of these approaches in the corresponding noise oblivious setting as defined in Section~\ref{sec:algorithms}. (NO-reRLS and NO-cGA). 

We measure the performance of each procedure by the number of calls to the objective function until the true optimum $1^n$ is generated. This performance metric is standard in the field of evolutionary computation because typically objective function evaluation is the most costly operation in terms of computation time.
For the cGA, this is twice the number of iterations through the while loop in Algorithm~\ref{alg:cga}. For reRLS, this is the number of iteration times the number of resamples necessary to obtain a suitable estimate of the true objective function value.

\begin{figure}
  \centering
  \begin{tikzpicture}
    \begin{loglogaxis}[%
      width=\columnwidth, 
      height=\columnwidth, 
      xlabel = $\sigma^2$, 
      ylabel = number of evaluations of $f$, 
      legend pos = south east, 
      legend style={draw=none},
      legend cell align=left,
      ]
      
      \addplot+[%
      name path=A1,
      no markers,
      forget plot,
      style={dotted},
      black,
      ]
      table[x=x,y=lq]{experiments/Boxplots/n=100RLSFPBoxplot.txt};     

      \addplot+[%
      name path=B1,
      no markers,
      forget plot,
      style={dotted},
      black
      ]
      table[x=x,y=uq]{experiments/Boxplots/n=100RLSFPBoxplot.txt};   
      
      \addplot[%
      color1!10,
      forget plot
      ] 
      fill between[of=A1 and B1];     

      \addplot+[%
      mark=x,
      color1,
      mark size=3
      ]
      table[x=x,y=med]{experiments/Boxplots/n=100RLSFPBoxplot.txt};    
      
      \addplot+[%
      name path=A2,
      no markers,
      forget plot,
      style={dotted},
      black
      ]
      table[x=x,y=lq] {experiments/Boxplots/n=100NoiseObliviousRLSFPBoxplot.txt};          

      \addplot+[%
      name path=B2,
      no markers,forget plot,
      style={dotted},
      black
      ]
      table[x=x,y=uq]{experiments/Boxplots/n=100NoiseObliviousRLSFPBoxplot.txt};    

      \addplot[%
      color2!10,
      forget plot
      ] fill between[of=A2 and B2];      

      \addplot+[%
      mark=+,
      color2,
      mark size=3
      ]
      table[x=x,y=med]{experiments/Boxplots/n=100NoiseObliviousRLSFPBoxplot.txt};    

      \addplot+[%
      name path=A3,
      no markers,
      forget plot,
      style={dotted},
      black
      ]
      table[x=x,y=lq]{experiments/Boxplots/n=100CGAFPBoxplot.txt};   
      
      \addplot+[%
      name path=B3,
      no markers,
      forget plot,
      style={dotted},
      black
      ]
      table[x=x,y=uq]{experiments/Boxplots/n=100CGAFPBoxplot.txt};   

      \addplot[
      color3!10,
      forget plot
      ] 
      fill between[of=A3 and B3];     

      \addplot+[%
      mark=o,
      color3
      ]
      table[x=x,y=med]{experiments/Boxplots/n=100CGAFPBoxplot.txt};    
      
      \addplot+[%
      name path=A4,
      no markers,
      forget plot,
      style={dotted},
      black
      ]
      table[x=x,y=lq]{experiments/Boxplots/n=100NoiseObliviousCGAFPBoxplot.txt};    

      \addplot+[%
      name path=B4,
      no markers,
      forget plot,
      style={dotted},
      black
      ]
      table[x=x,y=uq]{experiments/Boxplots/n=100NoiseObliviousCGAFPBoxplot.txt};    

      \addplot[%
      color4!10, 
      forget plot
      ] 
      fill between[of=A4 and B4];                 
      \addplot+[%
      mark=*,
      color4
      ]
      table[x=x,y=med]{experiments/Boxplots/n=100NoiseObliviousCGAFPBoxplot.txt};    

      \draw[dashed] 
      ({axis cs:10,0}|-{rel axis cs:0,0}) -- ({axis cs:10,0}|-{rel axis cs:0,1});
      \coordinate (label) at ({axis cs:10,0}|-{rel axis cs:0,1}) {};
      \node[below right= 3mm and 0mm of label] {$\sigma^2 = \sqrt{n}$};

      \legend{reRLS, NO-reRLS, cGA, NO-cGA};
      
    \end{loglogaxis}
  \end{tikzpicture}
  \caption{\label{fig:var} Median run time as a function of noise variance for $n=100$, 100 runs at each point. Shaded area denotes IQR.}
\end{figure}
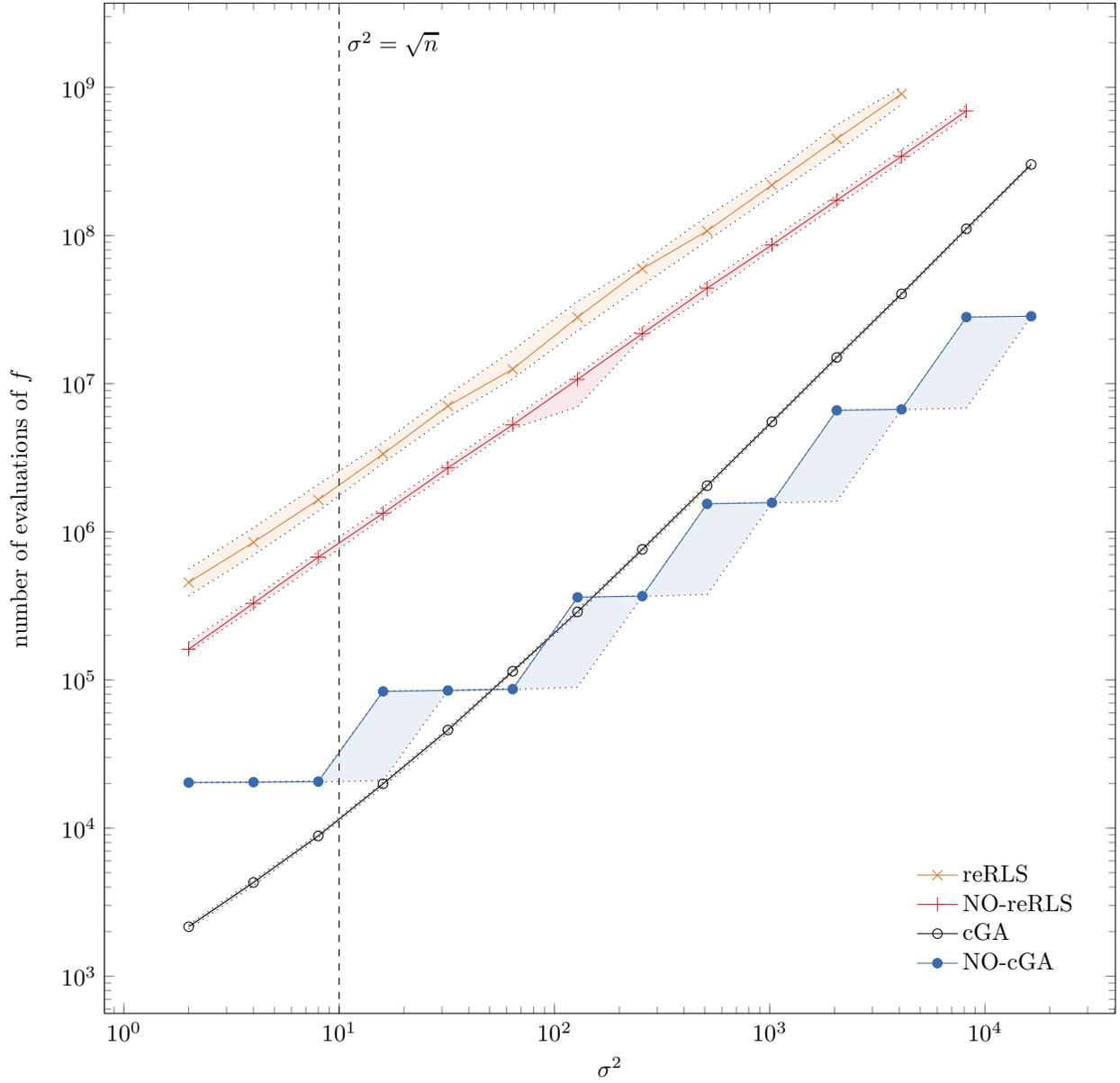

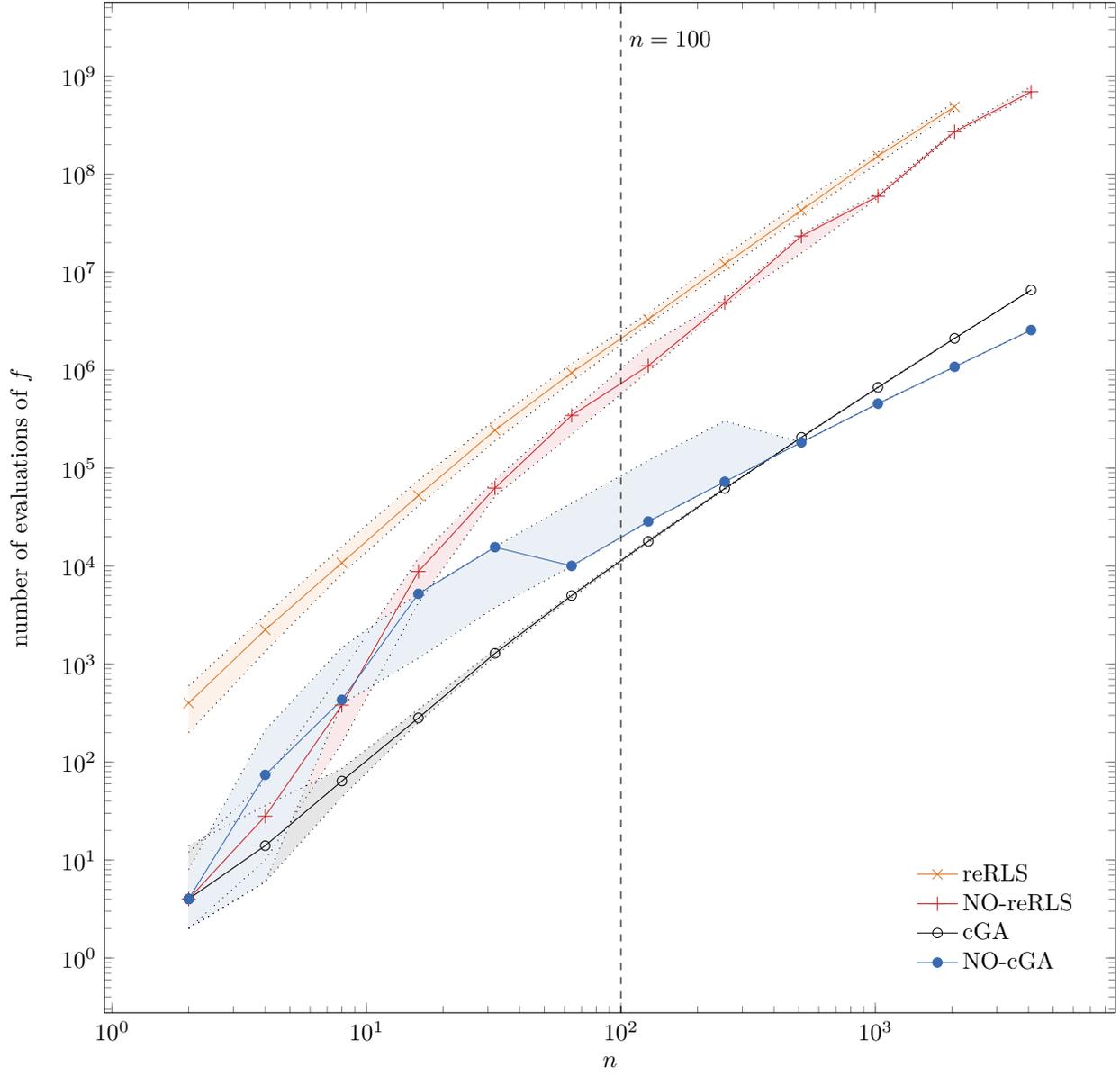
\begin{figure}
  \centering
  \begin{tikzpicture}

    \begin{loglogaxis}[%
      width=\columnwidth, 
      height=\columnwidth, 
      xlabel = $n$, 
      ylabel = number of evaluations of $f$,
      legend pos = south east, 
      legend style={draw=none},
      legend cell align=left
      ]
      
      \addplot+[%
      name path=C1,
      no markers,
      forget plot,
      style={dotted},
      black
      ]
      table[x=x,y=lq]{experiments/Boxplots/var=sqrt.n.RLSFPBoxplot.txt};    

      \addplot+[%
      name path=D1,
      no markers,
      forget plot,
      style={dotted},
      black
      ]
      table[x=x,y=uq]{experiments/Boxplots/var=sqrt.n.RLSFPBoxplot.txt};   

      \addplot+[%
      color1!10,
      forget plot
      ] fill between[of=C1 and D1];      

      \addplot+[%
      mark=x,
      color1,
      mark size=3
      ]
      table[x=x,y=med]{experiments/Boxplots/var=sqrt.n.RLSFPBoxplot.txt};    
      
      \addplot+[%
      name path=C2,
      no markers,
      forget plot,
      style={dotted},
      black
      ]
      table[x=x,y=lq]{experiments/Boxplots/var=sqrt.n.NoiseObliviousRLSFPBoxplot.txt};          
      \addplot+[%
      name path=D2,
      no markers,
      forget plot,
      style={dotted},
      black
      ]
      table[x=x,y=uq]{experiments/Boxplots/var=sqrt.n.NoiseObliviousRLSFPBoxplot.txt};    

      \addplot+[%
      color2!10,
      forget plot
      ] fill between[of=C2 and D2];      
      
      \addplot+[%
      mark=+,
      color2,
      mark size=3
     ]
     table[x=x,y=med]{experiments/Boxplots/var=sqrt.n.NoiseObliviousRLSFPBoxplot.txt};    

      \addplot+[%
      name path=C3,
      no markers,
      forget plot,
      style={dotted},
      black
      ]
      table[x=x,y=lq]{experiments/Boxplots/var=sqrt.n.CGAFPBoxplot.txt};   
      
      \addplot+[%
      name path=D3,
      no markers,
      forget plot,
      style={dotted},
      black
      ]
      table[x=x,y=uq]{experiments/Boxplots/var=sqrt.n.CGAFPBoxplot.txt};   

      \addplot+[%
      color3!10,
      forget plot
      ] 
      fill between[of=C3 and D3];     

      \addplot+[%
      mark=o,
      color3
      ]
      table[x=x,y=med]{experiments/Boxplots/var=sqrt.n.CGAFPBoxplot.txt};    
      
      \addplot+[%
      name path=C4,
      no markers,
      forget plot,
      style={dotted},
      black
      ]
      table[x=x,y=lq]{experiments/Boxplots/var=sqrt.n.NoiseObliviousCGAFPBoxplot.txt};    

      \addplot+[%
      name path=D4,
      no markers,
      forget plot,
      style={dotted},
      black
      ]
      table[x=x,y=uq]{experiments/Boxplots/var=sqrt.n.NoiseObliviousCGAFPBoxplot.txt};    

      \addplot+[%
      color4!10, 
      forget plot
      ] fill between[of=C4 and D4];                 

      \addplot+[%
      mark=*,
      color4
      ]
      table[x=x,y=med]{experiments/Boxplots/var=sqrt.n.NoiseObliviousCGAFPBoxplot.txt};    

      \draw[dashed] 
      ({axis cs:100,0}|-{rel axis cs:0,0}) -- ({axis cs:100,0}|-{rel axis cs:0,1});
      \coordinate (label) at ({axis cs:100,0}|-{rel axis cs:0,1}) {};
      \node[below right= 3mm and 0mm of label] {$n=100$};
      
      \legend{reRLS, NO-reRLS, cGA, NO-cGA};
    \end{loglogaxis}
  \end{tikzpicture}
  \caption{\label{fig:n} Median run time as a function of $n$ for $\sigma^2=\sqrt{n}$, 100 runs at each point. Shaded area denotes IQR.}

\end{figure}

The performance of each algorithm is plotted fixing $n=100$ and controlling the variance in Figure~\ref{fig:var}. For each procedure and variance value we run each algorithm 100 times until the true optimum is found and collect the number of calls to the objective function for each run. The median run times and their interquartile ranges are plotted. We also plot the performance as a function of $n$ (fixing $\sigma^2=\sqrt{n}$) in Figure~\ref{fig:n}. Both results are plotted on log-log plots; Thus the cGA variants are an order of magnitude faster than the baseline.

\onlyARXIV
{
Figures~\ref{fig:nK} and \ref{fig:varK} correspond to figures~\ref{fig:n} and \ref{fig:var}, respectively, and depict the number of re-evaluations ((NO-)reRLS) per iteration or the value
of $K$ ((NO-)cGA) that was sufficient for the respective algorithm to succeed. Note that the functions for the non-noise-oblivious algorithms have deterministic function values whereas
the ones for the noise-oblivious versions are random variables.
}

\onlyARXIV
{
\begin{figure}
  \centering
  \begin{tikzpicture}
    \begin{loglogaxis}[%
      width=\columnwidth, 
      height=\columnwidth, 
      xlabel = $\sigma^2$, 
      ylabel = number of re-evaluations per iteration or $K$, 
      legend pos = south east, 
      legend style={draw=none},
      legend cell align=left,
      ]
      
      \addplot+[%
      name path=A1,
      no markers,
      forget plot,
      style={dotted},
      black,
      ]
      table[x=x,y=lq]{experiments/Results/n=100RLSFPK.txt};     

      \addplot+[%
      name path=B1,
      no markers,
      forget plot,
      style={dotted},
      black
      ]
      table[x=x,y=uq]{experiments/Results/n=100RLSFPK.txt};   
      
      \addplot[%
      color1!10,
      forget plot
      ] 
      fill between[of=A1 and B1];     

      \addplot+[%
      mark=x,
      color1,
      mark size=3
      ]
      table[x=x,y=med]{experiments/Results/n=100RLSFPK.txt};    
      
      \addplot+[%
      name path=A2,
      no markers,
      forget plot,
      style={dotted},
      black
      ]
      table[x=x,y=lq] {experiments/Results/n=100NoiseObliviousRLSFPK.txt};          

      \addplot+[%
      name path=B2,
      no markers,forget plot,
      style={dotted},
      black
      ]
      table[x=x,y=uq]{experiments/Results/n=100NoiseObliviousRLSFPK.txt};    

      \addplot[%
      color2!10,
      forget plot
      ] fill between[of=A2 and B2];      

      \addplot+[%
      mark=+,
      color2,
      mark size=3
      ]
      table[x=x,y=med]{experiments/Results/n=100NoiseObliviousRLSFPK.txt};    

      \addplot+[%
      name path=A3,
      no markers,
      forget plot,
      style={dotted},
      black
      ]
      table[x=x,y=lq]{experiments/Results/n=100CGAFPK.txt};   
      
      \addplot+[%
      name path=B3,
      no markers,
      forget plot,
      style={dotted},
      black
      ]
      table[x=x,y=uq]{experiments/Results/n=100CGAFPK.txt};   

      \addplot[
      color3!10,
      forget plot
      ] 
      fill between[of=A3 and B3];     

      \addplot+[%
      mark=o,
      color3
      ]
      table[x=x,y=med]{experiments/Results/n=100CGAFPK.txt};    
      
      \addplot+[%
      name path=A4,
      no markers,
      forget plot,
      style={dotted},
      black
      ]
      table[x=x,y=lq]{experiments/Results/n=100NoiseObliviousCGAFPK.txt};    

      \addplot+[%
      name path=B4,
      no markers,
      forget plot,
      style={dotted},
      black
      ]
      table[x=x,y=uq]{experiments/Results/n=100NoiseObliviousCGAFPK.txt};    

      \addplot[%
      color4!10, 
      forget plot
      ] 
      fill between[of=A4 and B4];                 
      \addplot+[%
      mark=*,
      color4
      ]
      table[x=x,y=med]{experiments/Results/n=100NoiseObliviousCGAFPK.txt};    

      \draw[dashed] 
      ({axis cs:10,0}|-{rel axis cs:0,0}) -- ({axis cs:10,0}|-{rel axis cs:0,1});
      \coordinate (label) at ({axis cs:10,0}|-{rel axis cs:0,1}) {};
      \node[below right= 3mm and 0mm of label] {$\sigma^2 = \sqrt{n}$};

      \legend{reRLS, NO-reRLS, cGA, NO-cGA};
      
    \end{loglogaxis}
  \end{tikzpicture}
  \caption{\label{fig:varK} Median of number of re-evaluations per iteration or median of $K$ as a function of noise variance for $n=100$, 100 runs at each point. Shaded area denotes IQR.}
\end{figure}
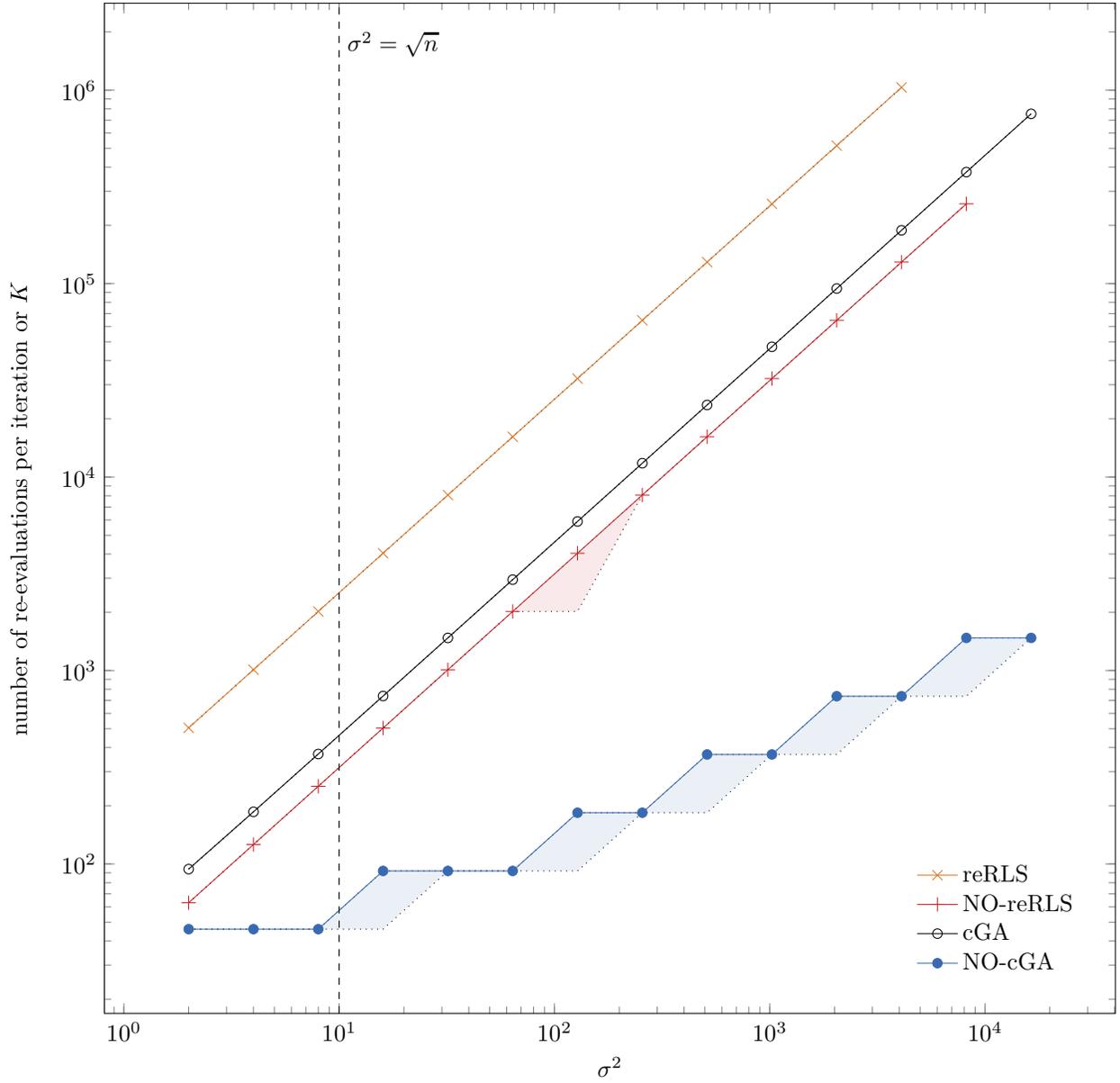

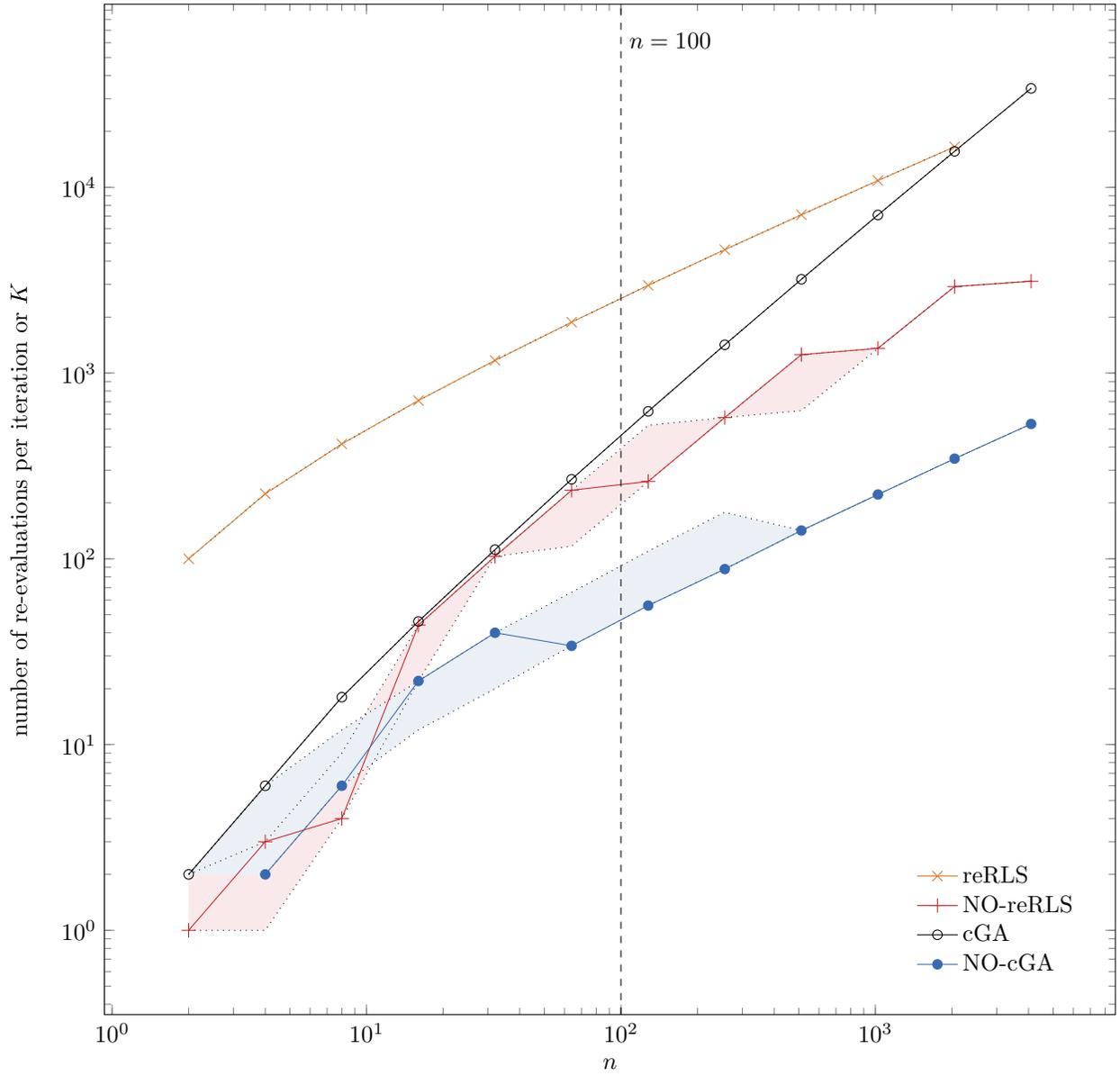
\begin{figure}
  \centering
  \begin{tikzpicture}

    \begin{loglogaxis}[%
      width=\columnwidth, 
      height=\columnwidth, 
      xlabel = $n$, 
      ylabel = number of re-evaluations per iteration or $K$,
      legend pos = south east, 
      legend style={draw=none},
      legend cell align=left
      ]
      
      \addplot+[%
      name path=C1,
      no markers,
      forget plot,
      style={dotted},
      black
      ]
      table[x=x,y=lq]{experiments/Results/var=sqrt.n.RLSFPK.txt};    

      \addplot+[%
      name path=D1,
      no markers,
      forget plot,
      style={dotted},
      black
      ]
      table[x=x,y=uq]{experiments/Results/var=sqrt.n.RLSFPK.txt};   

      \addplot+[%
      color1!10,
      forget plot
      ] fill between[of=C1 and D1];      

      \addplot+[%
      mark=x,
      color1,
      mark size=3
      ]
      table[x=x,y=med]{experiments/Results/var=sqrt.n.RLSFPK.txt};    
      
      \addplot+[%
      name path=C2,
      no markers,
      forget plot,
      style={dotted},
      black
      ]
      table[x=x,y=lq]{experiments/Results/var=sqrt.n.NoiseObliviousRLSFPK.txt};          
      \addplot+[%
      name path=D2,
      no markers,
      forget plot,
      style={dotted},
      black
      ]
      table[x=x,y=uq]{experiments/Results/var=sqrt.n.NoiseObliviousRLSFPK.txt};    

      \addplot+[%
      color2!10,
      forget plot
      ] fill between[of=C2 and D2];      
      
      \addplot+[%
      mark=+,
      color2,
      mark size=3
     ]
     table[x=x,y=med]{experiments/Results/var=sqrt.n.NoiseObliviousRLSFPK.txt};    

      \addplot+[%
      name path=C3,
      no markers,
      forget plot,
      style={dotted},
      black
      ]
      table[x=x,y=lq]{experiments/Results/var=sqrt.n.CGAFPK.txt};   
      
      \addplot+[%
      name path=D3,
      no markers,
      forget plot,
      style={dotted},
      black
      ]
      table[x=x,y=uq]{experiments/Results/var=sqrt.n.CGAFPK.txt};   

      \addplot+[%
      color3!10,
      forget plot
      ] 
      fill between[of=C3 and D3];     

      \addplot+[%
      mark=o,
      color3
      ]
      table[x=x,y=med]{experiments/Results/var=sqrt.n.CGAFPK.txt};    
      
      \addplot+[%
      name path=C4,
      no markers,
      forget plot,
      style={dotted},
      black
      ]
      table[x=x,y=lq]{experiments/Results/var=sqrt.n.NoiseObliviousCGAFPK.txt};    

      \addplot+[%
      name path=D4,
      no markers,
      forget plot,
      style={dotted},
      black
      ]
      table[x=x,y=uq]{experiments/Results/var=sqrt.n.NoiseObliviousCGAFPK.txt};    

      \addplot+[%
      color4!10, 
      forget plot
      ] fill between[of=C4 and D4];                 

      \addplot+[%
      mark=*,
      color4
      ]
      table[x=x,y=med]{experiments/Results/var=sqrt.n.NoiseObliviousCGAFPK.txt};    

      \draw[dashed] 
      ({axis cs:100,0}|-{rel axis cs:0,0}) -- ({axis cs:100,0}|-{rel axis cs:0,1});
      \coordinate (label) at ({axis cs:100,0}|-{rel axis cs:0,1}) {};
      \node[below right= 3mm and 0mm of label] {$n=100$};
      
      \legend{reRLS, NO-reRLS, cGA, NO-cGA};
    \end{loglogaxis}
  \end{tikzpicture}
  \caption{\label{fig:nK} Median of number of re-evaluations per iteration or median of $K$ as a function of $n$ for $\sigma^2=\sqrt{n}$, 100 runs at each point. Shaded area denotes IQR.}

\end{figure}
}


\onlyIJCAI{%
  \section{Discussion}
  \input{summary}
}

\onlyARXIV{%
  \section{Conclusions}
  In this paper we have examined the benefit of sexual recombination in evolutionary optimization in an uncertain environment. We introduce the concept of an algorithm \emph{scaling gracefully} with noise. We rigorously proved that mutation-only evolutionary algorithms do not scale gracefully in the sense that they cannot optimize noisy functions in polynomial time when the noise intensity is sufficiently high. On the other hand, we proved that a simple estimation of distribution algorithm that uses gene pool recombination can always optimize noisy OneMax ($\fnoise$) in polynomial time, subject only to the condition that the noise variance $\sigma^2$ is bounded by some polynomial in $n$. 

A common way to handle noisy objective functions is to modify the optimization algorithm to perform resampling in order to estimate the true value of the underlying objective function. We have also presented empirical results that show the sexual recombination algorithm optimizes $\fnoise$ an order of magnitude faster than a resampling hillclimber.
Our results highlight the importance of understanding the influence of different search operators in uncertain environments, and suggest that algorithms such as the compact genetic algorithm that use sexual recombination are able to scale gracefully with noise.


}
\clearpage 

\onlyIJCAI{
  \bibliographystyle{named}
}
\onlyARXIV{
  \bibliographystyle{plain}
}
\bibliography{../bibli,../references}

\begin{thebibliography}{10}

\bibitem{AroraHK12}
Sanjeev Arora, Elad Hazan, and Satyen Kale.
\newblock The multiplicative weights update method: a meta-algorithm and
  applications.
\newblock {\em Theory of Computing}, 8(1):121--164, 2012.

\bibitem{BackFZ97}
Thomas B{\"{a}}ck, David~B. Fogel, and Zbigniew Michalewicz, editors.
\newblock {\em Handbook of Evolutionary Computation}.
\newblock IOP Publishing Ltd., 1st edition, 1997.

\bibitem{Barton2014diverse}
Nicholas~H. Barton, Sebastian Novak, and Tiago Paixão.
\newblock Diverse forms of selection in evolution and computer science.
\newblock {\em Proceedings of the National Academy of Sciences},
  111(29):10398--10399, July 2014.

\bibitem{Bia-Dor-Gam-Gut:j:09}
L.~Bianchi, M.~Dorigo, L.~Gambardella, and W.~Gutjahr.
\newblock A {S}urvey on {M}etaheuristics for {S}tochastic {C}ombinatorial
  optimization.
\newblock {\em Natural Computing}, 8:239--287, 2009.

\bibitem{Chastain2014algorithms}
Erick Chastain, Adi Livnat, Christos Papadimitriou, and Umesh Vazirani.
\newblock Algorithms, games, and evolution.
\newblock {\em Proceedings of the National Academy of Sciences},
  111(29):10620--10623, July 2014.

\bibitem{Doe-Hot-Koe:c:12}
B.~Doerr, A.~Hota, and T.~K\"{o}tzing.
\newblock {A}nts {E}asily {S}olve {S}tochastic {S}hortest {P}ath {P}roblems.
\newblock In {\em Proc.~of GECCO'12}, pages 17--24, 2012.

\bibitem{Doerr2013adaptive}
Benjamin Doerr and Leslie~Ann Goldberg.
\newblock Adaptive drift analysis.
\newblock {\em Algorithmica}, 65(1):224--250, 2013.

\bibitem{DoerrHK12}
Benjamin Doerr, Edda Happ, and Christian Klein.
\newblock Crossover can provably be useful in evolutionary computation.
\newblock {\em Theor.\ Comput.\ Sci.}, 425:17--33, 2012.

\bibitem{Droste2006cga}
Stefan Droste.
\newblock A rigorous analysis of the compact genetic algorithm for linear
  functions.
\newblock {\em Natural Computing}, 5(3):257--283, August 2006.

\bibitem{EibenSmith03}
Agoston~E. Eiben and J.~E. Smith.
\newblock {\em Introduction to Evolutionary Computing}.
\newblock Springer, 2003.

\bibitem{Fel-Koe:c:13}
Matthias Feldmann and Timo K\"{o}tzing.
\newblock {O}ptimizing {E}xpected {P}ath {L}engths with {A}nt {C}olony
  {O}ptimization {U}sing {F}itness {P}roportional {U}pdate.
\newblock In {\em Proc.~of FOGA'13}, pages 65--74, 2013.

\bibitem{Goldberg1989}
David~E. Goldberg.
\newblock {\em Genetic Algorithms in Search, Optimization and Machine
  Learning}.
\newblock Addison-Wesley, 1989.

\bibitem{Gutjahr96}
W.~Gutjahr and G.~Pflug.
\newblock {S}imulated {A}nnealing for {N}oisy {C}ost {F}unctions.
\newblock {\em Journal of Global Optimization}, 8:1--13, 1996.

\bibitem{Harik1999cga}
Georges~R. Harik, Fernando~G. Lobo, and David~E. Goldberg.
\newblock The compact genetic algorithm.
\newblock {\em IEEE Trans. on Evol. Comp.}, 3(4):287--297, 1999.

\bibitem{JansenWegener:j:02}
Thomas Jansen and Ingo Wegener.
\newblock The analysis of evolutionary algorithms -- a proof that crossover
  really can help.
\newblock {\em Algorithmica}, 34(1):47--66, 2002.

\bibitem{Jansen2005c}
Thomas Jansen and Ingo Wegener.
\newblock Real royal road functions---where crossover provably is essential.
\newblock {\em Discrete Appl.\ Math.}, 149:111--125, 2005.

\bibitem{JinBranke:j:05:robustSurvey}
Yaochu Jin and J{\"u}rgen Branke.
\newblock Evolutionary optimization in uncertain environments---a survey.
\newblock {\em IEEE Trans. on Evol. Comp.}, 9:303--317, 2005.

\bibitem{Koetzing2014concentration}
Timo K\"{o}tzing.
\newblock Concentration of first hitting times under additive drift.
\newblock In {\em Proc.~of GECCO'14}, pages 1391--1397, 2014.

\bibitem{KotzingST:c:11:crossover}
Timo K{\"o}tzing, Dirk Sudholt, and Madeleine Theile.
\newblock How crossover helps in pseudo-boolean optimization.
\newblock In {\em Proc.\ of GECCO~'11}, pages 989--996, 2011.

\bibitem{Muehlenbein1996recombination}
Heinz M\"uhlenbein and Gerhard Paa\ss.
\newblock From recombination of genes to the estimation of distributions {I}.
  {B}inary parameters.
\newblock In {\em Proc. of PPSN IV}, pages 178--187. Springer-Verlag, 1996.

\bibitem{Muehlenbein1996genepool}
Heinz M\"uhlenbein and Hans-Michael Voigt.
\newblock Gene pool recombination in genetic algorithms.
\newblock In {\em Meta-Heuristics}, pages 53--62. Springer US, 1996.

\bibitem{NeumannWitt:b:10}
F.~Neumann and C.~Witt.
\newblock {\em Bioinspired Computation in Combinatorial Optimization --
  Algorithms and Their Computational Complexity.}
\newblock Springer, 2010.

\bibitem{NeumannORS:c:11}
Frank Neumann, Pietro~S. Oliveto, G{\"u}nter Rudolph, and Dirk Sudholt.
\newblock On the effectiveness of crossover for migration in parallel
  evolutionary algorithms.
\newblock In {\em Proc.~of GECCO'11}, pages 1587--1594, 2011.

\bibitem{Oliveto2012erratum}
P.~S. Oliveto and C.~Witt.
\newblock Erratum: Simplified drift analysis for proving lower bounds in
  evolutionary computation.
\newblock {\em {ArXiv} e-prints}, 2012.

\bibitem{Oliveto2011simplified}
Pietro~S. Oliveto and Carsten Witt.
\newblock Simplified drift analysis for proving lower bounds in evolutionary
  computation.
\newblock {\em Algorithmica}, 59(3):369--386, 2011.

\bibitem{PrugelBennett:j:10}
Adam Pr{\"u}gel-Bennett.
\newblock {B}enefits of a {P}opulation: {F}ive {M}echanisms {T}hat {A}dvantage
  {P}opulation-{B}ased {A}lgorithms.
\newblock {\em IEEE Trans.\ on Evol.\ Comp.}, 14:500--517, 2010.

\bibitem{Richter2008}
J.~Neal Richter, Alden Wright, and John Paxton.
\newblock Ignoble trails - where crossover is provably harmful.
\newblock In {\em Proc.\ of PPSN~'08}, pages 92--101, 2008.

\bibitem{Storch2004}
Tobias Storch and Ingo Wegener.
\newblock Real royal road functions for constant population size.
\newblock {\em Theor.\ Comput.\ Sci.}, 320:123--134, 2004.

\bibitem{Sud-Thy:j:12}
D.~Sudholt and C.~Thyssen.
\newblock A {S}imple {A}nt {C}olony {O}ptimizer for {S}tochastic {S}hortest
  {P}ath problems.
\newblock {\em Algorithmica}, 64:643--672, 2012.

\bibitem{Sudholt2005}
Dirk Sudholt.
\newblock Crossover is provably essential for the {Ising} model on trees.
\newblock In {\em Proc.\ of GECCO~'05}, pages 1161--1167, 2005.

\bibitem{Teytaud2015noisy}
Olivier Teytaud.
\newblock Private communication, 2015.

\bibitem{vonBahr1965inequalities}
Bengt von Bahr and Carl-Gustav Esseen.
\newblock Inequalities for the $r$-th absolute moment of a sum of random
  variables, $1 \leq r \leq 2$.
\newblock {\em The Annals of Mathematical Statistics}, 36(1):299--303, 1965.

\bibitem{Watson2007}
Richard~A. Watson and Thomas Jansen.
\newblock A building-block royal road where crossover is provably essential.
\newblock In {\em Proc.\ of GECCO~'07}, pages 1452--1459, 2007.

\bibitem{wolfram:erfc}
Eric~W Weisstein.
\newblock Erfc, 2015.
\newblock From MathWorld--A Wolfram Web Resource.
  \url{http://mathworld.wolfram.com/Erfc.html}.

\end{thebibliography}

\end{document}